\documentclass[journal,twoside,web]{ieeecolor2}
\usepackage{generic}
\usepackage{cite}
\usepackage{amsmath,amssymb,amsfonts}
\usepackage[boxed,ruled]{algorithm2e}
\usepackage{graphicx}
\usepackage{multirow}
\usepackage{textcomp}
\usepackage{balance}

\newcommand{\xq}[0]{\textcolor{black}}
\def\BibTeX{{\rm B\kern-.05em{\sc i\kern-.025em b}\kern-.08em
    T\kern-.1667em\lower.7ex\hbox{E}\kern-.125emX}}
\begin{document}
\title{FreqDINO++: A Frequency-Guided Multi-Task Routing Vision Foundation Model for \protect\\ Universal Ultrasound Analysis}
\author{Qing Xu, Yixuan Zhang, Yue Li, Xiangjian He, \IEEEmembership{Senior Member, IEEE}, Qian Zhang, Mainul Haque,  \\Rong Qu, \IEEEmembership{Fellow, IEEE}, Wenting Duan, Jieyun Bai, Zhen Chen
\thanks{This work is partially supported by the Zhejiang Department of Transportation General Research and Development Project (2024039), National Natural Science Foundation of China grant (62476037), Ningbo Science and Technology Bureau under NBNSF General Programme (2025J114), and PolyU Start-up Fund (P0060371).\textit{(Equal contribution: Q.
Xu, Y. Zhang and Y. Li, Corresponding author: X. He and Z. Chen)}}
\thanks{\quad Q. Xu, Y. Zhang, L. Yue, X. He, and Q. Zhang are with School of Computer Science, University of Nottingham Ningbo China, Ningbo, Zhejiang, China (e-mail: sean.he@nottingham.edu.cn).}
\thanks{\quad M. Haque is with School of Mathematical Sciences, University of Nottingham Ningbo China, Ningbo, Zhejiang, China (e-mail: Mainul.Haque@nottingham.edu.cn).}
\thanks{\quad R. Qu is with the School of Computer Science, University of Nottingham, Nottingham NG72RD, UK (email: rong.qu@nottingham.ac.uk).}
\thanks{\quad W. Duan is with School of Engineering and Physical Science, University of Lincoln, Lincoln  LN6 7TS, UK (email: wduan@lincoln.ac.uk).}
\thanks{\quad J. Bai is with College of Information Science and Technology, Jinan University, Guangzhou, China (email: jbai996@aucklanduni.ac.nz).}
\thanks{\quad Z. Chen is with Department of Data Science and Artificial Intelligence, The Hong Kong Polytechnic University, Hong Kong SAR. (e-mail: z.chen@polyu.edu.hk).}}

\maketitle

\begin{abstract}
Ultrasound image analysis plays a crucial role in cancer screening and prenatal diagnosis, yet comprehensive assessment requires jointly addressing tasks such as lesion segmentation and benign-malignant classification. While recent vision foundation models have shown remarkable universal representations, unlocking their potential for ultrasound is bottlenecked by the considerable domain gap from natural images. Existing methods typically fine-tune heavy vision encoders for isolated tasks, incurring substantial computational overhead while overlooking the underlying commonalities across heterogeneous tasks. In this work, we propose FreqDINO++, a frequency-guided multi-task routing vision foundation model for universal ultrasound analysis. We first introduce a Multi-task Routing Adapter (MR-Adapter) to support parameter-efficient integration of task-common and task-specific knowledge, a Frequency-aware Feature Enhancer (F$^2$-Enhancer) is then designed to capture the rich multi-scale frequency characteristics of ultrasound images, and a Task-aligned Collaborative Decoder (TC-Decoder) is devised to promote collaboration between dense and global prediction tasks through global-local token interaction. Extensive experiments on large-scale multi-task and external single-task ultrasound benchmarks demonstrate that FreqDINO++ consistently outperforms strong baselines and recent foundation models across 27 diverse clinical task scenarios, while also showing promising generalization to unseen data. The code is at https://github.com/MingLang-FD/FreqDINO-Plus.
\end{abstract}


\section{Introduction}
\label{sec:introduction}
Ultrasound image analysis plays a crucial role in clinical practice and has received significant research attention owing to its real-time, non-invasive, and cost-effective imaging characteristics \cite{chen2022recent,jiao2024usfm,lin2025deep}. For example, breast ultrasound examination requires precise lesion segmentation for volumetric estimation, diagnostic classification for benign-malignant differentiation, and bounding box localization for multi-lesion detection \cite{chen2022aau,tao2025enhancing,qin2025real}. Similarly, obstetric ultrasound relies on anatomical structure segmentation for organ delineation, fetal standard plane classification for quality control, and keypoint regression for biometric measurements such as femur length and angle of progression \cite{chen2024multi,mikolaj2025predicting}. These demands of heterogeneous tasks have given rise to the challenging field of universal ultrasound image analysis, which seeks to unify segmentation, classification, detection, and regression within a single framework to serve diverse clinical workflows.

\begin{figure*}[!t]
  \centering
  \includegraphics[width=1\linewidth]{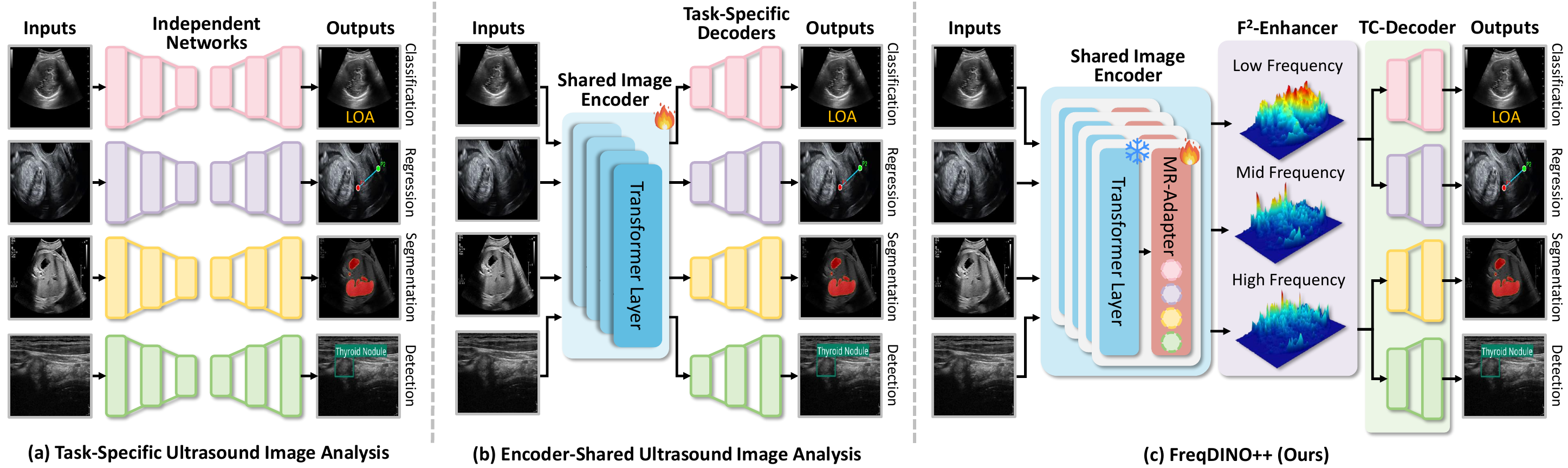}
\caption{Comparison of our FreqDINO++ and existing ultrasound image analysis works. (a) Independent task-specific networks separately handle heterogeneous ultrasound tasks. (b) A shared vision encoder with task-specific decoders for heterogeneous ultrasound tasks. (c) Our FreqDINO++ leverages multi-task routing and frequency-guided enhancement with token-collaborative decoding for universal ultrasound analysis.}
  \label{fig:intro}
\end{figure*}

The classical ultrasound image analysis methods \cite{chen2022aau,jiao2024usfm,luo2025lgffm} have been designed for specific tasks, resulting in isolated solutions that address segmentation, classification, detection, and regression separately, as shown in Fig. \ref{fig:intro}(a). These conventional approaches typically adopt independent architectures without shared representations during training. For instance, segmentation networks \cite{ronneberger2015u,zhou2019unet++,isensee2021nnu,zhang2025annular,wang2025echo} are optimized to delineate tumor or organ boundaries at the pixel level for volumetric quantification, classification models \cite{wang2026deep,jiang2025segmentation,dosovitskiy2021an} differentiate benign-malignant lesions or recognize fetal standard planes for diagnostic support, detection frameworks \cite{carion2020end,tao2025enhancing,qin2025real} localize multiple lesions with bounding boxes for early screening, and regression methods \cite{zhu2025machine,mikolaj2025predicting} estimate anatomical landmarks such as head circumference or femur length for biometric assessment. However, maintaining multiple task-specific networks \cite{schlemper2019attention,ibtehaz2023acc} leads to substantial computational overhead and deployment burden in clinical settings, motivating the development of a unified framework with shared representations that could serve multiple ultrasound analysis tasks simultaneously.

To this end, Vision Foundation Models (VFMs) pre-trained on large-scale datasets have emerged as a promising backbone for ultrasound image analysis. In particular, the recent DINOv3 \cite{simeoni2025dinov3} has demonstrated remarkable representation capacity through self-supervised learning, offering high-quality features for diverse downstream tasks. However, existing methods \cite{chen2022aau, chen2024transunet, nam2024modality} typically rely on fully fine-tuning the heavy pre-trained encoder for individual task, which is computationally costly and hinders multi-task deployment in clinical practice, as shown in Fig. \ref{fig:intro}(b). Moreover, these methods typically treat each task in isolation and underexploit the underlying commonalities across heterogeneous tasks, where segmentation and detection share dense prediction cues such as boundaries and spatial localization, while classification and regression rely on similar global semantic context. More critically, because DINOv3 is pre-trained on natural images, its representations do not explicitly account for the low contrast, speckle noise, and frequency-dependent textures characteristic of ultrasound images \cite{nam2024modality,gao2025dino,yang2025segdino}. These observations motivate us to develop a unified framework that efficiently adapts VFM to ultrasound imaging while coordinating heterogeneous downstream tasks for comprehensive clinical analysis.

To overcome these limitations, we propose FreqDINO++, a frequency-guided multi-task routing vision foundation model for universal ultrasound analysis. As illustrated in Fig. \ref{fig:intro}(c), FreqDINO++ adapts the frozen DINOv3 backbone to ultrasound scenarios through task-conditioned routing and frequency-guided enhancement within a unified framework. Specifically, we devise a Multi-Task Routing Adapter (MR-Adapter) to seamlessly unify task-common and task-specific knowledge for parameter-efficient multi-task adaptation. We further design a Frequency-aware Feature Enhancer (F$^2$-Enhancer) to fully exploit the multi-scale frequency characteristics of ultrasound images for discriminative feature representations. Moreover, we devise a Task-aligned Collaborative Decoder (TC-Decoder) that harnesses the interaction between global and local token representations to coordinate heterogeneous downstream tasks. Extensive experiments on diverse ultrasound datasets demonstrate that FreqDINO++ consistently outperforms state-of-the-art methods across all task categories.

The contributions of this work are summarized as follows:
\begin{itemize}
\item We propose FreqDINO++, a frequency-guided multi-task routing vision foundation model that adapts DINOv3 to universal ultrasound analysis, unifying heterogeneous clinical tasks within a single framework.
\item We design the MR-Adapter coupled with F$^2$-Enhancer that seamlessly unifies task-common and task-specific knowledge through task-conditioned routing, while fully exploiting the multi-scale frequency characteristics inherent in ultrasound images for discriminative feature representations.
\item We devise the TC-Decoder that harnesses the collaborative interaction between global and local token representations to coordinate heterogeneous downstream tasks, facilitating complementary knowledge transfer across dense and global prediction objectives.
\item We validate FreqDINO++ on large-scale multi-task and external benchmark datasets, where it demonstrates strong performance across 27 clinical task scenarios and promising generalization to unseen data.
\end{itemize}

A preliminary version of this work has been published in ISBI 2026 \cite{zhang2025freqdino}. In this paper, we have made substantial extensions with the following highlights: 1) We extend the scope from single-task segmentation to universal ultrasound analysis that jointly addresses segmentation, classification, detection, and regression; 2) We devise the MR-Adapter that replaces the generic lightweight adapter in the conference work \cite{zhang2025freqdino}, integrating task-common and task-specific knowledge through task-conditioned routing; 3) We reformulate the segmentation-oriented MFEA and FGBR into a unified F$^2$-Enhancer, extending frequency modeling from one segmentation task to 27 diverse clinical scenarios with multi-scale frequency characteristics; 4) We upgrade the dual-head boundary-guided decoder to a TC-Decoder that coordinates heterogeneous tasks via global-local token interaction; 5) We conduct extensive experiments on five datasets including the large-scale FMC-UIA challenge spanning 27 clinical task scenarios, together with comprehensive ablation studies.

\begin{figure*}[!t]
  \centering
  \includegraphics[width=1\linewidth]{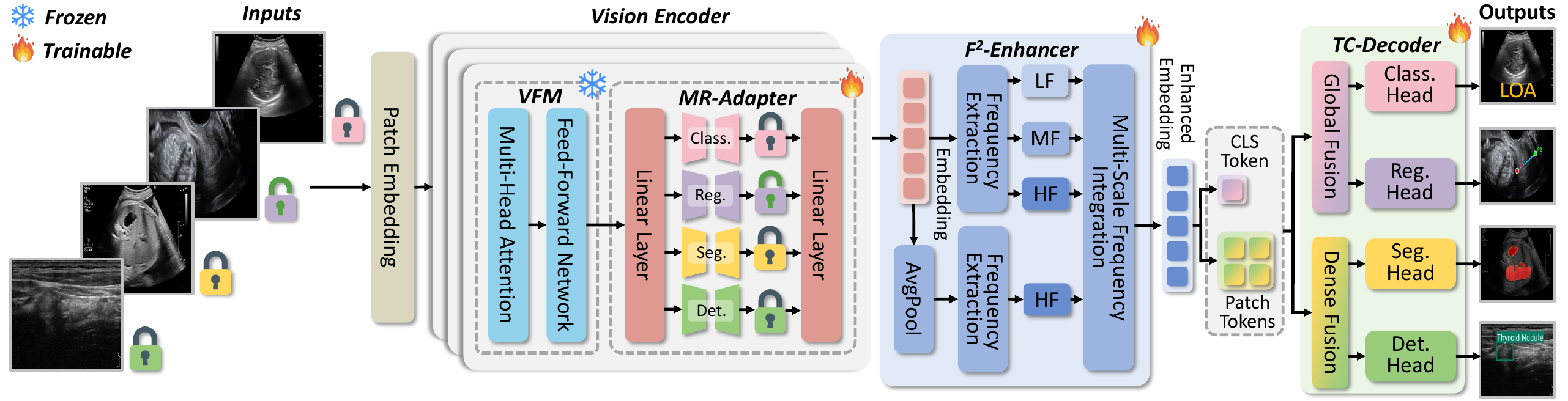}
\caption{The overview of our FreqDINO++ framework for universal ultrasound analysis, consisting of an MR-Adapter that integrates task-common and task-specific knowledge through task-conditioned routing, an F$^2$-Enhancer that captures multi-scale frequency characteristics inherent in ultrasound images, and a TC-Decoder with global and dense heads for coordinating heterogeneous downstream tasks. The FreqDINO++ framework efficiently adapts the VFM to ultrasound scenarios and coordinates heterogeneous tasks within a unified architecture.}
  \label{fig:framework}
\end{figure*}

\section{Related Work}
\subsection{Ultrasound Image Analysis}

Ultrasound imaging has become one of the most widely used modalities in clinical practice, owing to its real-time capability, non-invasive nature, and cost-effectiveness \cite{chen2022recent,lin2025deep}. In routine clinical workflows, ultrasound examination often requires the joint execution of multiple analytical tasks to support comprehensive patient assessment. For instance, breast ultrasound relies on precise lesion segmentation for volumetric estimation, diagnostic classification for benign-malignant differentiation, and bounding box localization for multi-lesion detection \cite{chen2022aau,tao2025enhancing}. Similarly, obstetric ultrasound demands anatomical structure segmentation for organ delineation, standard plane classification for quality control, and keypoint regression for biometric measurements \cite{chen2024multi,mikolaj2025predicting}. However, ultrasound images inherently suffer from speckle noise, low contrast, and frequency-dependent texture variations \cite{nam2024modality,luo2025lgffm}, posing significant challenges for automated analysis. These heterogeneous task demands and imaging characteristics have motivated the development of advanced computational methods for universal ultrasound image analysis.

In this context, deep learning methods have made notable strides across individual ultrasound analysis tasks through tailored network architectures. For segmentation, U-Net \cite{ronneberger2015u} and its variants \cite{zhou2019unet++,isensee2021nnu,ibtehaz2023acc} have served as the predominant architectures, with subsequent works introducing attention mechanisms \cite{schlemper2019attention,chen2022aau}, transformer-based designs \cite{chen2024transunet}, and frequency-domain fusion strategies \cite{luo2025lgffm}. For classification and detection, vision transformers \cite{dosovitskiy2021an,jiang2025segmentation} and DETR-based frameworks \cite{carion2020end,tao2025enhancing,qin2025real} have been adopted for diagnostic categorization and lesion localization, respectively. Despite the progress, these methods are predominantly designed for specific tasks in isolation, requiring separate optimization and deployment pipelines and incurring substantial computational overhead while neglecting the commonalities across heterogeneous tasks. In contrast, the proposed FreqDINO++ framework unifies heterogeneous ultrasound tasks within a single architecture, enabling universal analysis without maintaining separate task-specific networks.

\subsection{Vision Foundation Models in Medical Imaging}
 
The vision foundation models pre-trained on large-scale datasets have emerged as a promising paradigm for medical image analysis. The SAM series \cite{kirillov2023segment,ravisam} revolutionized segmentation through interactive prompting mechanisms, enabling remarkable generalization across diverse visual domains. Subsequent adaptations have tailored SAM to medical imaging, including MedSAM \cite{ma2024segment} for universal medical segmentation, Medical SAM Adapter \cite{wu2025medical} for parameter-efficient domain adaptation, and SAM2-Adapter \cite{chen2025sam2} for extending SAM\,2 to broader downstream tasks. \xq{In the ultrasound domain, USFM \cite{jiao2024usfm} constructed a multi-organ database containing over two million images and employed spatial-frequency dual masked image modeling to learn universal ultrasound representations for segmentation and classification. However, USFM \cite{jiao2024usfm} captures only ultrasound-specific knowledge through domain-exclusive pre-training and treats each downstream task in isolation by attaching independent task-specific heads, overlooking the commonalities across heterogeneous tasks. In contrast, our proposed FreqDINO++ framework retains the general visual knowledge from the frozen DINOv3 backbone and injects ultrasound-specific knowledge through parameter-efficient adaptation, while coordinating heterogeneous tasks through collaborative decoding.}
 
Recent advances have also further explored prompt-driven and encoder-based adaptations specifically for ultrasound analysis. CC-SAM \cite{gowda2024cc} incorporated cross-feature attention to handle low-contrast boundaries in ultrasound images, Lin \emph{et al.} \cite{lin2024beyond} designed an auto-prompting mechanism for end-to-end ultrasound segmentation, Zhang \emph{et al.} \cite{zhang2025adapting} adapted vision foundation models for real-time ultrasound analysis, and Chen \emph{et al.} \cite{chen2024multi} proposed a multi-organ foundation model with task prompts and anatomical priors. Meanwhile, DINOv3 \cite{simeoni2025dinov3} has demonstrated superior dense feature quality through self-supervised learning, inspiring adaptations such as DINO U-Net \cite{gao2025dino}, SegDINO \cite{yang2025segdino}, and GuiDINO \cite{liang2026guidino} that exploit its high-fidelity representations for medical segmentation. Despite the progress, these methods remain constrained by parameter-heavy adaptation. Different from these approaches, FreqDINO++ adapts DINOv3 through task-conditioned routing and frequency-guided enhancement, achieving parameter-efficient multi-task adaptation that coordinates dense and global prediction objectives within a unified framework.

\section{Methods}

\subsection{Overview of FreqDINO++}
We present FreqDINO++ as a unified framework for universal ultrasound image analysis in Fig. \ref{fig:framework}. Given an ultrasound image from the $k$-th task, we first feed it into a frozen DINOv3 ViT-Large backbone, where each Transformer block is equipped with a MR-Adapter to perform task-conditioned routing. The adapted patch tokens are then processed by the F$^2$-Enhancer, which progressively aggregates multi-scale frequency information to enhance feature representations. Finally, the enhanced features along with the CLS token are delivered to the TC-Decoder to jointly serve segmentation, classification, detection, and regression tasks.

In general, our FreqDINO++ is specifically designed to address the challenges of adapting vision foundation models to multi-task ultrasound analysis. The MR-Adapter enables efficient task-conditioned routing at every encoder layer without maintaining separate adapters for each task. The F$^2$-Enhancer exploits the frequency-dependent characteristics of ultrasound images to enhance feature representations. Meanwhile, the TC-Decoder bridges dense and global tasks through token-level collaboration. Together, these modules enable FreqDINO++ to achieve universal ultrasound analysis within a single unified framework.

\begin{figure}[!t]
  \centering
  \includegraphics[width=1\linewidth]{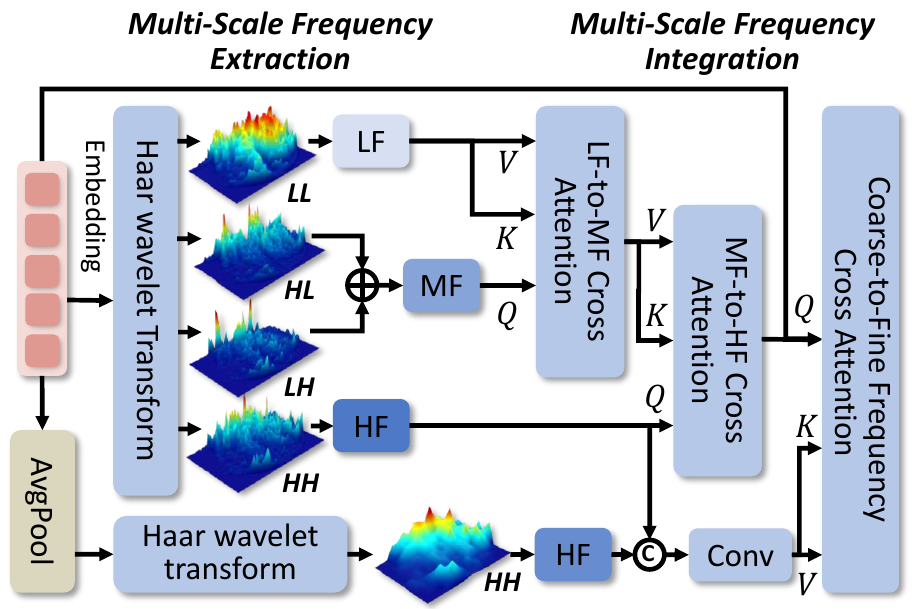}
    \caption{The illustration of F$^2$-Enhancer, decomposing adapted features into multiple frequency bands at two complementary scales via parallel Haar wavelet transforms and progressively integrating them through cascaded cross-band attention, providing frequency-enhanced representations for diverse downstream tasks.}
  \label{fig:m1}
\end{figure}

\subsection{Multi-Task Routing Adapter}
While DINOv3 \cite{simeoni2025dinov3} delivers powerful general-purpose representations, fully fine-tuning its heavy vision encoder for every ultrasound task is computationally prohibitive and quickly becomes intractable as the number of tasks grows. \xq{The conventional adapters \cite{chen2022vision,jie2023revisiting} alleviate this cost by inserting lightweight modules into the frozen backbone, but they learn a single shared set of parameters across all tasks, failing to capture task-specific variations for heterogeneous ultrasound analysis.} To address this, we devise the MR-Adapter that efficiently integrates task-common with task-specific knowledge within a single adapter module, enabling flexible multi-task adaptation in a parameter-efficient manner.

Specifically, the MR-Adapter is inserted after the feed-forward network in each Transformer block and employs a shared bottleneck architecture with deterministic task-conditioned routing. Given the input token sequence $\mathbf{x} \in \mathbb{R}^{B \times N \times C}$ and a task identifier $t \in \{\text{seg}, \text{cls}, \text{det}, \text{reg}\}$, the MR-Adapter first projects the input into a low-dimensional bottleneck space of dimension $r$ through a shared down-projection $\mathbf{W}_{\text{down}} \in \mathbb{R}^{C \times r}$. Within this compact space, $T$ lightweight task-specific transformation pathways $\{\mathbf{E}_t\}_{t=1}^{T}$ are maintained, each implemented as a MLP layer that captures task-specific adaptation patterns. The task identifier $t$ deterministically selects the corresponding pathway $\mathbf{E}_t$ without any learnable gating or routing probability, ensuring stable training free from expert collapse. The transformed features are then mapped back to the original dimension via a shared up-projection $\mathbf{W}_{\text{up}} \in \mathbb{R}^{r \times C}$. This process is formulated as:
\begin{equation} \label{eq:mra}
\mathbf{y} = \mathbf{x} + \mathbf{W}_{\text{up}} \cdot \sigma\!\Big(\mathbf{E}_t\big(\mathbf{W}_{\text{down}} \cdot \mathbf{x}\big)\Big),
\end{equation}
where $\sigma(\cdot)$ denotes the GELU activation function applied after the task-specific routing. Since $\mathbf{W}_{\text{down}}$ and $\mathbf{W}_{\text{up}}$ are shared across all tasks, the MR-Adapter naturally aligns feature spaces of different tasks while the lightweight pathways $\{\mathbf{E}_t\}$ introduce only minimal task-specific parameters, achieving substantial parameter savings compared with maintaining independent per-task adapters. Through this design, the MR-Adapter achieves fine-grained task-aware adaptation from the earliest encoder layers, enabling effective cross-task knowledge sharing while preserving task-specific specialization for universal ultrasound analysis.

\begin{figure}[!t]
  \centering
  \includegraphics[width=1\linewidth]{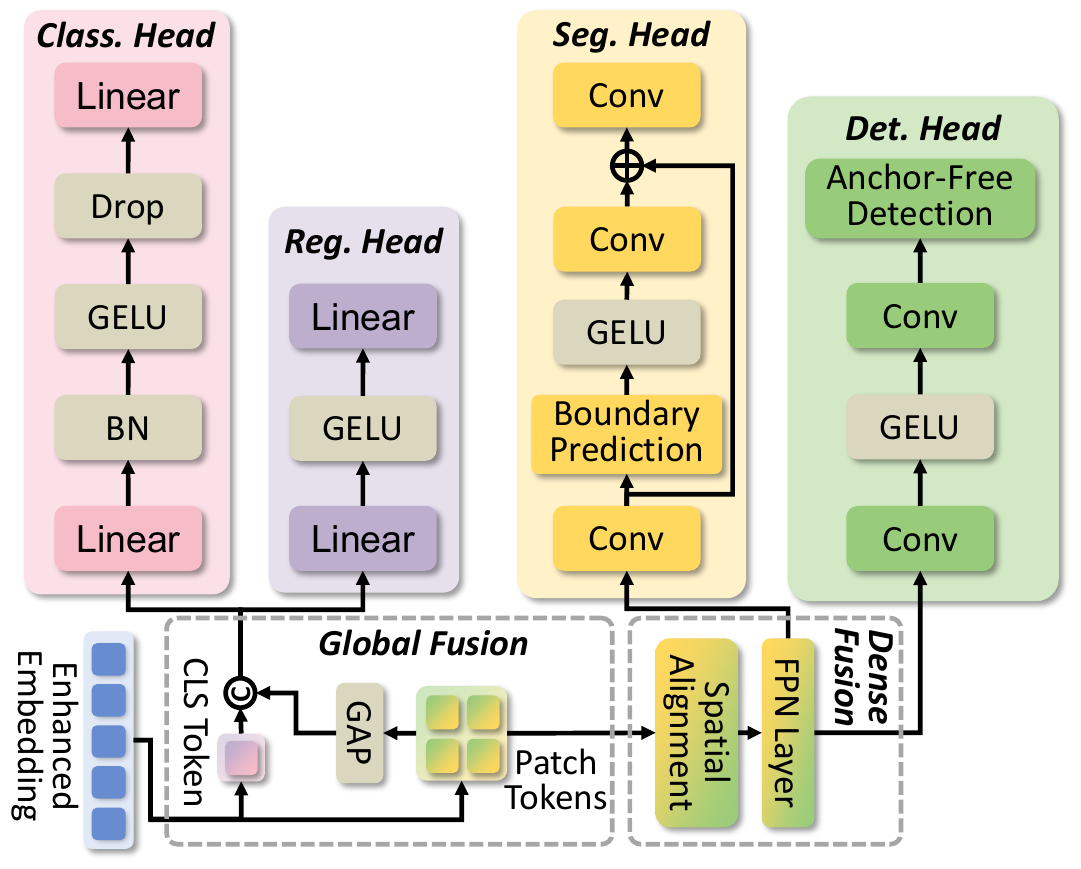}
    \caption{The architecture of the TC-Decoder, which exploits the intrinsic alignment between task objectives and representation granularity: classification and regression are coupled through a global fusion pathway driven by semantic representations, while segmentation and detection are coupled through a dense fusion pathway driven by spatial representations, achieving harmonious multi-task collaboration.}
  \label{fig:m2}
\end{figure}

\subsection{Frequency-Aware Feature Enhancer}
Ultrasound images encode clinically significant information across distinct frequency bands: high-frequency components capture boundary details critical for segmentation and detection, mid-frequency components characterize tissue textures essential for classification, and low-frequency components describe global anatomical structures that underpin regression-based biometric measurements \cite{zhang2025freqdino}. \xq{Existing frequency-domain methods \cite{nam2024modality,luo2025lgffm} perform frequency decomposition at a single spatial scale, which limits their ability to simultaneously capture fine-grained boundary details and coarse structural patterns across different receptive fields.} However, the DINOv3 backbone, operating entirely in the spatial domain, cannot explicitly exploit such frequency-dependent characteristics. To this end, we propose the F$^2$-Enhancer that fully exploits the multi-scale frequency characteristics inherent in ultrasound images to produce discriminative feature representations for diverse downstream tasks, as shown in Fig. \ref{fig:m1}.

\begin{algorithm}[t]
\caption{Optimization Pipeline for FreqDINO++}
\label{alg:freqdino}
\KwIn{Ultrasound datasets $\{\mathcal{D}_t\}_{t=1}^{T}$; Frozen DINOv3 encoder $E$; MR-Adapter $\{\mathbf{W}_{\text{down}}, \mathbf{W}_{\text{up}}, \{\mathbf{E}_t\}_{t=1}^{T}\}$; F$^2$-Enhancer $\mathcal{E}$; TC-Decoder $\mathcal{D}_{\text{TC}}$}
\KwOut{Trained FreqDINO++ model}
\For{epoch $= 1$ to $N_{\text{epoch}}$}{
  Construct unified index pool $\mathcal{P} = \bigcup_{t=1}^{T} \mathcal{P}_t$\;
  \While{$\mathcal{P}$ is not exhausted}{
    Sample task $t \sim \text{Uniform}(\{1, \dots, T\})$\;
    Sample a task-homogeneous mini-batch $\mathcal{B}_t$ from $\mathcal{P}_t$\;
    \For{each $\mathbf{x} \in \mathcal{B}_t$}{
      $\mathbf{F}_{\text{patch}}, \mathbf{t}_{\text{cls}} = E(\mathbf{x};\, t)$ via Eq. \eqref{eq:mra}\;
      Reshape $\mathbf{F}_{\text{patch}}$ to $\mathbf{F} \in \mathbb{R}^{C \times H \times W}$\;
      $\mathbf{F}_{\text{enh}} = \mathcal{E}(\mathbf{F})$ via Eq. \eqref{eq:freq_bands}-\eqref{eq:freq_fuse}\;
    }
    \eIf{$t \in \{\text{seg}, \text{det}\}$}{
      $\hat{y} = \mathcal{D}_{\text{TC}}^{\text{local}}(\mathbf{F}_{\text{enh}})$\;
    }{
      $\hat{y} = \mathcal{D}_{\text{TC}}^{\text{global}}(\mathbf{F}_{\text{enh}}, \mathbf{t}_{\text{cls}})$ via Eq. \eqref{eq:cls_fuse}-\eqref{eq:global_heads}\;
    }
    Compute $\mathcal{L}_{\text{FreqDINO}}$ and update parameters\;
  }
}
\end{algorithm}

Particularly, given the spatial feature map $\mathbf{F} \in \mathbb{R}^{B \times C \times H \times W}$ reshaped from the adapted patch tokens, we apply two parallel Haar wavelet transforms operating at the original and a coarser scale to extract multi-scale frequency information. The original-scale branch directly decomposes $\mathbf{F}$ into four subbands $\{\mathbf{F}_{LL}, \mathbf{F}_{LH}, \mathbf{F}_{HL}, \mathbf{F}_{HH}\}$ and groups them into three frequency bands as follows:
\begin{equation} \label{eq:freq_bands}
\begin{aligned}
\mathbf{F}_{\text{LF}} &= \phi_L(\mathbf{F}_{LL}), \\
\mathbf{F}_{\text{MF}} &= \phi_M(\mathbf{F}_{LH} + \mathbf{F}_{HL}), \\
\mathbf{F}_{\text{HF}} &= \phi_H(\mathbf{F}_{HH}),
\end{aligned}
\end{equation}
where $\phi_L$, $\phi_M$, and $\phi_H$ are $1{\times}1$ convolutions for channel alignment, and the mid-frequency band is obtained by element-wise summation of the two directional detail subbands. In parallel, the coarse-scale branch first applies an average pooling operation $\mathbf{F}^{\downarrow} = \text{AvgPool}(\mathbf{F})$ and then performs Haar wavelet decomposition, retaining only the HH subband as $\mathbf{F}_{\text{HF}}^{c} = \phi_H^{c}(\mathbf{F}_{HH}^{\downarrow})$ to provide compact high-frequency structural cues over an enlarged receptive field, while discarding the coarse LL, LH, and HL subbands to avoid redundancy with the low- and mid-frequency context already preserved in the fine-scale branch. To enable effective cross-band interaction, we flatten each band into token sequences and design a cascaded cross-attention mechanism that progressively propagates contextual information from low to high frequencies. The mid-frequency tokens first attend to the low-frequency tokens to absorb global structural context, and the enriched mid-frequency tokens then serve as context for the high-frequency tokens:
\begin{equation} \label{eq:cross_attn}
\begin{aligned}
\hat{\mathbf{Z}}_{\text{MF}} &= \mathbf{Z}_{\text{MF}} + \text{CA}(\mathbf{Z}_{\text{MF}},\, \mathbf{Z}_{\text{LF}}), \\
\hat{\mathbf{Z}}_{\text{HF}} &= \mathbf{Z}_{\text{HF}} + \text{CA}(\mathbf{Z}_{\text{HF}},\, \hat{\mathbf{Z}}_{\text{MF}}),
\end{aligned}
\end{equation}
where $\text{CA}(\mathbf{Q}, \mathbf{K})$ denotes the multi-head cross-attention with $\mathbf{Q}$ as query and $\mathbf{K}$ as both key and value. To further integrate the multi-scale boundary cues, we concatenate the enriched fine-scale and the coarse-scale high-frequency tokens and apply a $1{\times}1$ convolution to produce the fused multi-scale value, upon which a coarse-to-fine cross-attention is performed with the enriched fine-scale tokens as the query:
\begin{equation} \label{eq:cf_attn}
\begin{aligned}
\mathbf{V}_{\text{cf}} &= \text{Conv}\big(\text{Cat}[\hat{\mathbf{Z}}_{\text{HF}},\, \mathbf{Z}_{\text{HF}}^{c}]\big), \\
\hat{\mathbf{Z}}_{\text{cf}} &= \hat{\mathbf{Z}}_{\text{HF}} + \text{CA}(\hat{\mathbf{Z}}_{\text{HF}},\, \mathbf{V}_{\text{cf}}).
\end{aligned}
\end{equation}
The integrated representation is then projected back to the spatial domain with a residual connection:
\begin{equation} \label{eq:freq_fuse}
\mathbf{F}_{\text{enh}} = \mathbf{F} + \lambda \cdot \text{Proj}(\hat{\mathbf{Z}}_{\text{cf}}),
\end{equation}
where $\lambda$ is a learnable scaling factor. On this basis, our proposed F$^2$-Enhancer module explicitly captures multi-scale frequency information inherent in ultrasound images, enabling the enhanced representations to simultaneously encode boundary details, tissue textures, and anatomical structures that collectively benefit all downstream tasks.

\subsection{Task-Aligned Collaborative Decoder}
\xq{Existing ultrasound analysis methods typically establish independent decoders for each task \cite{chen2022aau,tao2025enhancing,mikolaj2025predicting,chen2024multi,lin2024beyond}, where dense prediction tasks (segmentation and detection) and global prediction tasks (classification and regression) are handled by entirely separate architectures without any feature interaction.} This isolation not only introduces substantial computational redundancy but also prevents complementary knowledge transfer between related tasks, such as boundary-aware features benefiting both segmentation and detection, or global context supporting both classification and regression. To overcome this limitation, we design the TC-Decoder that leverages the collaborative interaction between local and global token representations to jointly serve all four tasks within a unified architecture, as shown in Fig. \ref{fig:m2}.

Specifically, the TC-Decoder receives the frequency-enhanced spatial features $\mathbf{F}_{\text{enh}}$ and the CLS token $\mathbf{t}_{\text{cls}} \in \mathbb{R}^{B \times C}$ from the encoder, and routes them into two complementary branches through a Dense Fusion module and a Global Fusion module, respectively. For the local branch serving dense prediction tasks, the Dense Fusion module first applies a Spatial Alignment operation that transforms the single-scale patch tokens into multi-scale feature pyramids with strides $\{4, 8, 16, 32\}$, followed by a Feature Pyramid Network (FPN) layer for hierarchical feature aggregation. \xq{The aggregated features are then delivered to a boundary-guided segmentation head that first predicts boundary maps and fuses them back into the main feature stream via residual summation before the final mask prediction, and an anchor-free detection head following the CenterNet \cite{duan2019centernet}, which consists of three parallel prediction branches: a center-point heatmap branch supervised by a penalty-reduced pixel-wise focal loss, a size regression branch predicting width and height vectors with L1 loss, and a local offset branch compensating for discretization errors with L1 loss, where inference extracts top-K peaks via $3\times3$ max pooling without requiring Non-Maximum Suppression.} For the global branch serving holistic prediction tasks, the Global Fusion module fuses the spatial and semantic representations through CLS token-guided feature fusion:
\begin{equation} \label{eq:cls_fuse}
\mathbf{h}_{\text{fused}} = \text{Cat}\big(\text{GAP}(\mathbf{F}_{\text{enh}}),\, \mathbf{t}_{\text{cls}}\big) \in \mathbb{R}^{B \times 2C},
\end{equation}
where $\text{GAP}(\cdot)$ denotes global average pooling that compresses the spatial features into a compact global representation, and $\text{Cat}(\cdot)$ concatenates it with the CLS token to form a joint embedding that encodes both frequency-enhanced spatial information and high-level semantic knowledge. The fused representation is then fed into a classification head and a regression head as follows:
\begin{equation} \label{eq:global_heads}
\begin{aligned}
\hat{y}_{\text{cls}} &= \mathbf{W}_c \cdot \text{Dropout}\big(\text{BN}(\mathbf{h}_{\text{fused}})\big), \\
\hat{y}_{\text{reg}} &= \mathbf{W}_r \cdot \mathbf{h}_{\text{fused}},
\end{aligned}
\end{equation}
where $\mathbf{W}_c$ and $\mathbf{W}_r$ are linear projection weights, with the classification branch further applying batch normalization (BN) and dropout for regularization, while the regression branch directly projects the fused features to preserve coordinate precision. In this way, our TC-Decoder enables dense and global tasks to share the same frequency-enhanced representations while maintaining task-appropriate decoding pathways, facilitating complementary knowledge transfer across heterogeneous tasks for universal ultrasound analysis.

\subsection{Training and Optimization Pipeline}
We summarize the training and optimization pipeline of FreqDINO++ for universal ultrasound analysis in Algorithm~\ref{alg:freqdino}. To construct our framework, we initialize the frozen DINOv3 encoder $E$, the MR-Adapter parameters $\{\mathbf{W}_{\text{down}}, \mathbf{W}_{\text{up}}, \{\mathbf{E}_t\}_{t=1}^{T}\}$, the F$^2$-Enhancer $\mathcal{F}_{\text{F$^2$-Enhancer}}$, and the TC-Decoder $\mathcal{D}_{\text{TC}}$. The framework adopts a task-homogeneous batch training strategy, where each mini-batch contains data from only a single task to avoid gradient conflicts caused by heterogeneous loss scales, while a uniform task sampler ensures balanced training across all tasks within each epoch.

For each training iteration, a task identifier $t$ is first uniformly sampled from the task pool, and a mini-batch $\mathcal{B}_t$ is drawn from the corresponding dataset $\mathcal{D}_t$. The input images are encoded through the frozen DINOv3 backbone with the MR-Adapter performing task-conditioned routing at each Transformer block. The resulting patch tokens are reshaped into spatial feature maps and enhanced by the F$^2$-Enhancer through cascaded wavelet decomposition and progressive cross-attention aggregation. The TC-Decoder then routes the enhanced features to the appropriate branch: the local branch for segmentation and detection, or the global branch for classification and regression.

The training is task-specific, with each task employing a tailored loss function. The overall objective of our FreqDINO++ framework can be formulated as follows:
\begin{equation} \label{eq:loss}
\begin{split}
\mathcal{L}_{\text{FreqDINO}} = \mathcal{L}_{\text{Dice}} + \lambda_{\text{seg}} \mathcal{L}_{\text{BCE}} + \lambda_{\text{cls}} \mathcal{L}_{\text{CE}} \\
+ \lambda_{\text{det}} \mathcal{L}_{\text{CenterNet}} + \lambda_{\text{reg}} \mathcal{L}_{\text{MSE}},
\end{split}
\end{equation}
where $\mathcal{L}_{\text{BCE}}$ denotes the binary cross-entropy loss computed on morphologically generated boundary labels, and $\lambda_{\text{seg}}$, $\lambda_{\text{cls}}$, $\lambda_{\text{det}}$, $\lambda_{\text{reg}}$ are weighting factors that balance the gradient magnitudes across tasks. Note that in each training iteration, only the loss terms corresponding to the sampled task $t$ are activated, while the remaining terms are set to zero. By alternating optimization across the four tasks under this pipeline, our FreqDINO++ framework achieves comprehensive ultrasound image analysis across segmentation, classification, detection, and regression.

\begin{table*}[!t]
\centering
\small
\setlength\tabcolsep{5.5pt}
\caption{Comparison with state-of-the-art methods on multi-task ultrasound analysis on the FMC-UIA dataset.}
\resizebox{\textwidth}{!}{\begin{tabular}{l|ccc|ccc|ccc|ccc}
\hline
\multirow{2}{*}{Methods} & \multicolumn{3}{c|}{Segmentation} & \multicolumn{3}{c|}{Classification} & \multicolumn{3}{c|}{Detection} & \multicolumn{3}{c}{Regression} \\
\cline{2-13}
& DSC$\uparrow$ & ASD$\downarrow$ & HD$\downarrow$ & AUC$\uparrow$ & F1$\uparrow$ & MCC$\uparrow$ & IoU$\uparrow$ & mAP$\uparrow$ & AP$_{\rm 50}$$\uparrow$ & MRE$\downarrow$ & MAE$\downarrow$ & SDR$\uparrow$ \\
\hline
AAU-Net \cite{chen2022aau} & 70.91 & 16.75 & 54.53 & 82.78 & 60.23 & 48.10 & 50.92 & 18.84 & 48.62 & 17.40 & 34.66 & 32.28 \\
TransUNet \cite{chen2024transunet} & 78.08 & 13.94 & 51.80 & 89.39 & 73.63 & 63.89 & 54.08 & 18.19 & 54.94 & 19.88 & 43.74 & 24.87 \\
MADGNet \cite{nam2024modality} & 75.14 & 16.12 & 56.56 & 89.94 & 65.88 & 55.21 & 57.59 & 24.30 & 65.32 & 18.48 & 31.92 & 32.46 \\
SAM2 \cite{ravisam} & 84.42 & 9.67 & 31.21 & 77.40 & 46.37 & 33.32 & 52.35 & 22.66 & 48.26 & 21.54 & 40.69 & 20.18 \\
Med-SA \cite{wu2025medical} & 88.43 & 7.37 & 21.75 & 90.22 & 78.63 & 70.13 & 56.90 & 20.78 & 63.98 & 26.61 & 59.12 & 16.08 \\
SAM2-Adapter \cite{chen2025sam2} & 84.23 & 9.95 & 32.33 & 88.50 & 62.47 & 52.80 & 53.27 & 20.52 & 47.14 & 14.96 & 27.16 & 41.12 \\
SAMUS \cite{lin2024beyond} & 80.67 & 13.21 & 49.17 & 90.18 & 75.43 & 67.66 & 62.71 & 31.90 & 64.09 & 22.40 & 40.13 & 29.51 \\
UltraSAM \cite{meyer2025ultrasam} & 87.42 & 7.89 & 24.48 & 89.77 & 68.95 & 58.73 & 55.76 & 20.46 & 55.95 & 18.37 & 37.60 & 24.99 \\
Nora \cite{wei2025noise} & 87.53 & 7.53 & 23.77 & 90.41 & 74.93 & 65.64 & 57.47 & 23.40 & 57.96 & 16.96 & 37.56 & 29.59 \\
USFM \cite{jiao2024usfm} & 86.96 & 7.62 & 24.25 & 87.85 & 73.30 & 63.18 & 46.78 & 22.93 & 48.13 & 12.33 & 23.93 & 54.57 \\
LGFFM \cite{luo2025lgffm} & 86.99 & 7.54 & 24.02 & 90.38 & 74.42 & 66.07 & 69.30 & 38.53 & 77.78 & 14.91 & 32.40 & 36.10 \\
DINOv3-FD \cite{he2026incentivizing} & 82.81 & 10.46 & 35.30 & 86.79 & 76.15 & 61.76 & 61.07 & 31.25 & 57.51 & 12.45 & 25.41 & 49.57 \\
\xq{UniSeg \cite{ye2023uniseg}} & \xq{85.71} & \xq{8.57} & \xq{26.21} & \xq{90.27} & \xq{80.59} & \xq{69.86} & \xq{66.29} & \xq{35.43} & \xq{67.34} & \xq{11.73} & \xq{21.72} & \xq{56.85} \\
\xq{Hermes \cite{gao2024training}} & \xq{86.19} & \xq{8.16} & \xq{24.18} & \xq{90.84} & \xq{79.11} & \xq{68.29} & \xq{66.49} & \xq{34.58} & \xq{63.18} & \xq{10.43} & \xq{19.21} & \xq{60.42} \\
\xq{TADFormer \cite{baek2025tadformer}} & \xq{86.46} & \xq{8.02} & \xq{23.88} & \xq{91.03} & \xq{78.50} & \xq{66.46} & \xq{68.69} & \xq{38.46} & \xq{76.46} & \xq{10.54} & \xq{19.96} & \xq{57.45} \\
\hline
FreqDINO++ & \textbf{88.46} & \textbf{6.62} & \textbf{20.32} & \textbf{91.24} & \textbf{81.57} & \textbf{71.99} & \textbf{74.83} & \textbf{47.75} & \textbf{82.84} & \textbf{7.74} & \textbf{14.21} & \textbf{71.69} \\
\hline
\end{tabular}}
\label{tab:multi_task}
\end{table*}

\begin{figure*}[!t]
  \centering
  \includegraphics[width=0.95\linewidth]{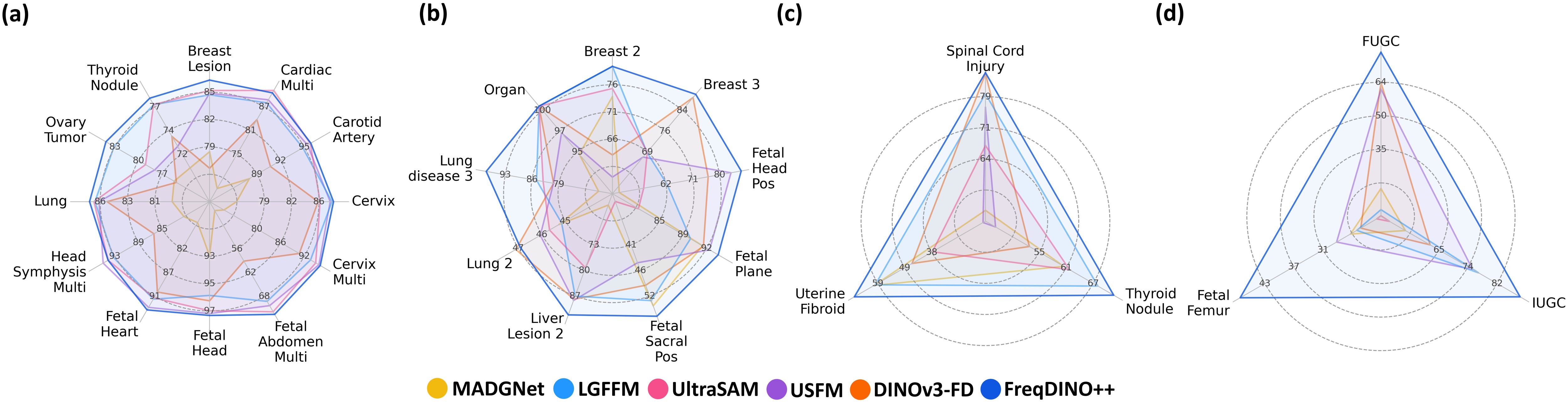}
  \caption{Per-subtask comparison of FreqDINO++ against five representative baselines on the FMC-UIA dataset across all 27 clinical task scenarios, organized by task category: (a) segmentation (DSC, 12 subtasks), (b) classification (F1, 9 subtasks), (c) detection (mAP, 3 subtasks), and (d) regression (SDR, 3 subtasks). FreqDINO++ consistently encloses the polygons of competing methods across diverse anatomies and tasks.}
  \label{fig:radar}
\end{figure*}

\section{Experiments}
\subsection{Datasets and Implementations}
\subsubsection{Datasets}
To validate the effectiveness of the proposed FreqDINO++, we conduct comprehensive evaluations across three experimental settings: (1) multi-task evaluation on a large-scale challenge dataset, (2) single-task evaluation on individual benchmark datasets, and (3) generalization evaluation on an unseen dataset. Specifically, we adopt the FMC-UIA 2026 challenge dataset \cite{lu2022jnu,chen2025comt,zhang2025automatic,chen2024psfhs} as the primary multi-task benchmark, which provides tens of thousands of ultrasound images from multi-center cohorts covering 27 subtasks across four task categories. For single-task evaluation, we employ BUSI \cite{al2020dataset} for segmentation and classification, TN5000 \cite{zhang2025tn5000} for detection, and FHC \cite{van2018automated} for regression. Moreover, we evaluate the generalization capability of FreqDINO++ on the unseen TN3K \cite{gong2023thyroid} dataset. The details are as follows:

\noindent \textbf{FMC-UIA} \cite{lu2022jnu,chen2025comt,zhang2025automatic,chen2024psfhs} is a large-scale multi-task ultrasound image analysis challenge dataset from multi-center cohorts, covering 27 subtasks across four task categories: segmentation (12 subtasks, 15,991 images) targeting structures such as fetal head, fetal lung, breast tumor, and thyroid nodule; classification (9 subtasks, 16,362 images) including standard plane recognition, tumor classification, and organ identification; detection (3 subtasks, 4,333 images) for thyroid nodule, uterine fibroid, and spinal cord; and regression (3 subtasks, 3,702 images) for angle of progression, cervical length, and fetal femur length measurement.

\noindent \textbf{BUSI} \cite{al2020dataset} is a breast ultrasound image dataset containing 780 images from 600 female patients, categorized into three classes: normal (133 images), benign (437 images), and malignant (210 images). Each image is provided with pixel-level segmentation masks, making it suitable for both segmentation and classification tasks.

\noindent \textbf{TN5000} \cite{zhang2025tn5000} is a large-scale open-access thyroid nodule ultrasound dataset comprising 5,000 B-mode images with bounding box annotations and biopsy-confirmed labels.
 
\noindent \textbf{FHC} \cite{van2018automated} is a fetal head circumference measurement dataset from the HC18 challenge, containing 1,334 two-dimensional ultrasound images annotated with ellipse fittings for head circumference estimation.

\noindent \textbf{TN3K} \cite{gong2023thyroid} is a thyroid nodule segmentation dataset containing 3,493 ultrasound images with pixel-level annotations. We use TN3K as an unseen dataset to evaluate the generalization capability of FreqDINO++ on the segmentation task.

\subsubsection{Implementation Details}
\label{sec:implementation}
We perform all experiments on four NVIDIA RTX 5880 Ada GPUs with 48GB using the PyTorch framework. \xq{To enable comparison under a unified multi-task setting, we equipped competing backbones with task-specific output heads following their original design principles, and trained them under matched optimization settings.} \xq{For SAM-based baselines (SAM2, Med-SA, SAM2-Adapter, SAMUS, and UltraSAM), no manual prompts are provided to the prompt encoder: the sparse prompt embeddings are set to empty, while the dense prompt embeddings default to the pre-trained no-mask embedding, ensuring all methods operate under the same prompt-free multi-task evaluation protocol.} We utilize the pre-trained DINOv3 ViT-Large \cite{simeoni2025dinov3} as the frozen backbone encoder of our FreqDINO++. We apply the AdamW optimizer with an initial learning rate of $1\times10^{-3}$ and a weight decay of 0.05, and employ the cosine annealing strategy to gradually decay the learning rate to $1\times10^{-6}$. The loss weighting factors in Eq. \eqref{eq:loss} are set as $\lambda_{\text{seg}}{=}0.3$, $\lambda_{\text{cls}}{=}1.0$, $\lambda_{\text{det}}{=}0.25$, and $\lambda_{\text{reg}}{=}50.0$ to balance the gradient magnitudes across different tasks. We split all datasets into the training, validation, and test sets as 8:1:1, and all input images are resized to $256 \times 256$. The batch size and the training epoch are set to 32 and 150, respectively. For data augmentation, the segmentation task employs horizontal flip, vertical flip, random brightness contrast, and Gaussian noise, while the other three tasks apply random brightness contrast and Gaussian noise.

\subsubsection{Evaluation Metrics}
To perform a comprehensive evaluation of universal ultrasound analysis, we adopt the following standard metrics for each task category. For segmentation, we employ the Dice Similarity Coefficient (DSC), Average Surface Distance (ASD), and Hausdorff Distance (HD) to assess region overlap, average boundary deviation, and maximum boundary error, respectively. For classification, we utilize the Area Under the ROC Curve (AUC), F1 Score, and Matthews Correlation Coefficient (MCC) to evaluate discriminative ability, precision-recall balance, and overall prediction quality. For detection, we adopt Intersection over Union (IoU), mean Average Precision (mAP), and Average Precision at IoU threshold of 0.5 (AP$_{\rm 50}$) to quantify localization accuracy and detection performance. For regression, we employ the Mean Radial Error (MRE), Mean Absolute Error (MAE), and Successful Detection Rate (SDR) to measure positional deviation and the proportion of predictions within a clinically acceptable distance threshold. In the comparison results, the best performance values are highlighted in \textbf{bold}.

\begin{figure*}[!t]
  \centering
  \includegraphics[width=1\linewidth]{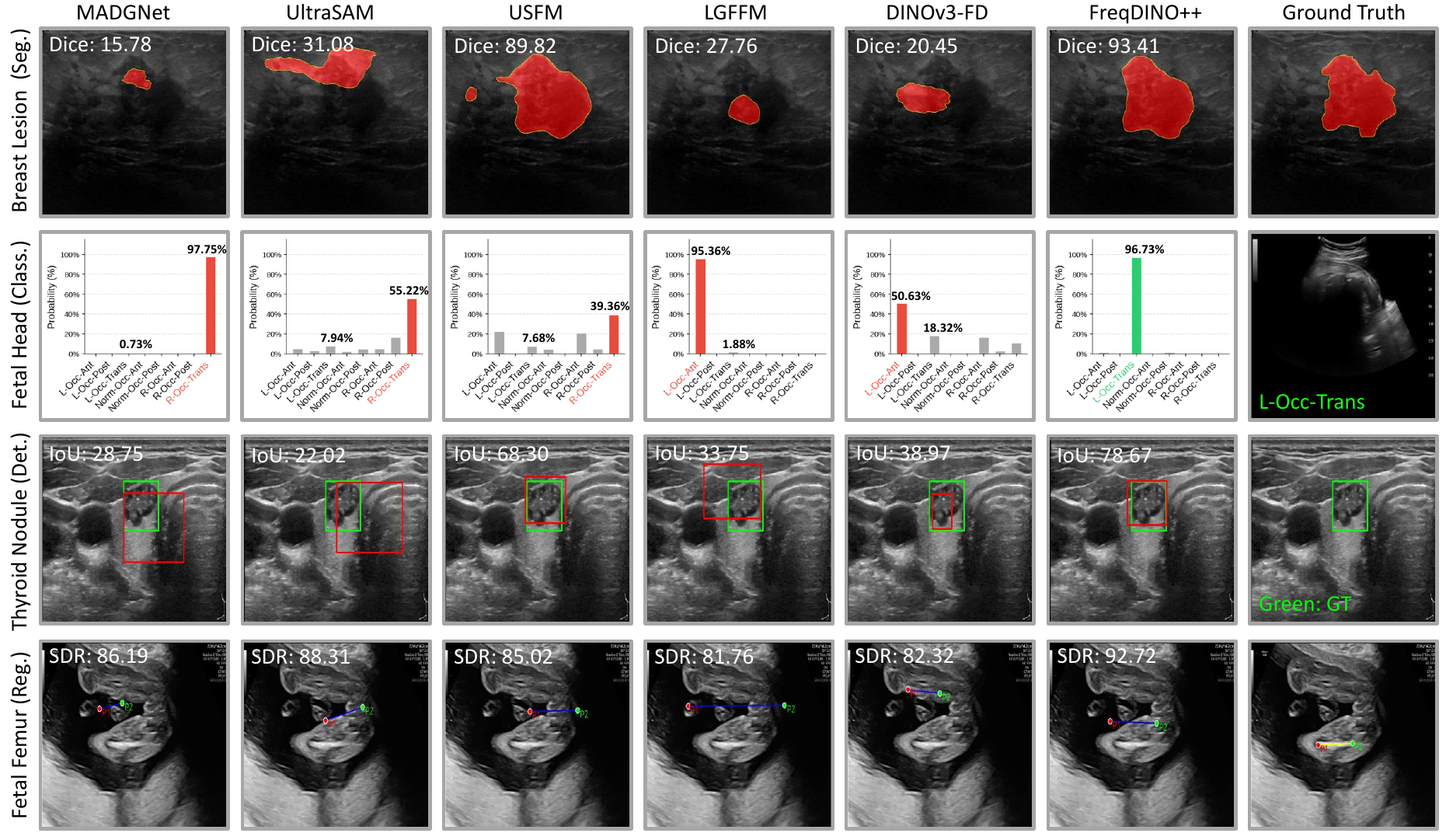}
  \caption{Visualization of qualitative results across four tasks on the FMC-UIA dataset. Our FreqDINO++ exhibits the best results, producing more accurate lesion boundaries, correct diagnostic categories with higher confidence, tighter detection boxes, and more precise biometric landmarks.}
  \label{fig:vis1}
\end{figure*}

\begin{figure}[!t]
  \centering
  \includegraphics[width=1\linewidth]{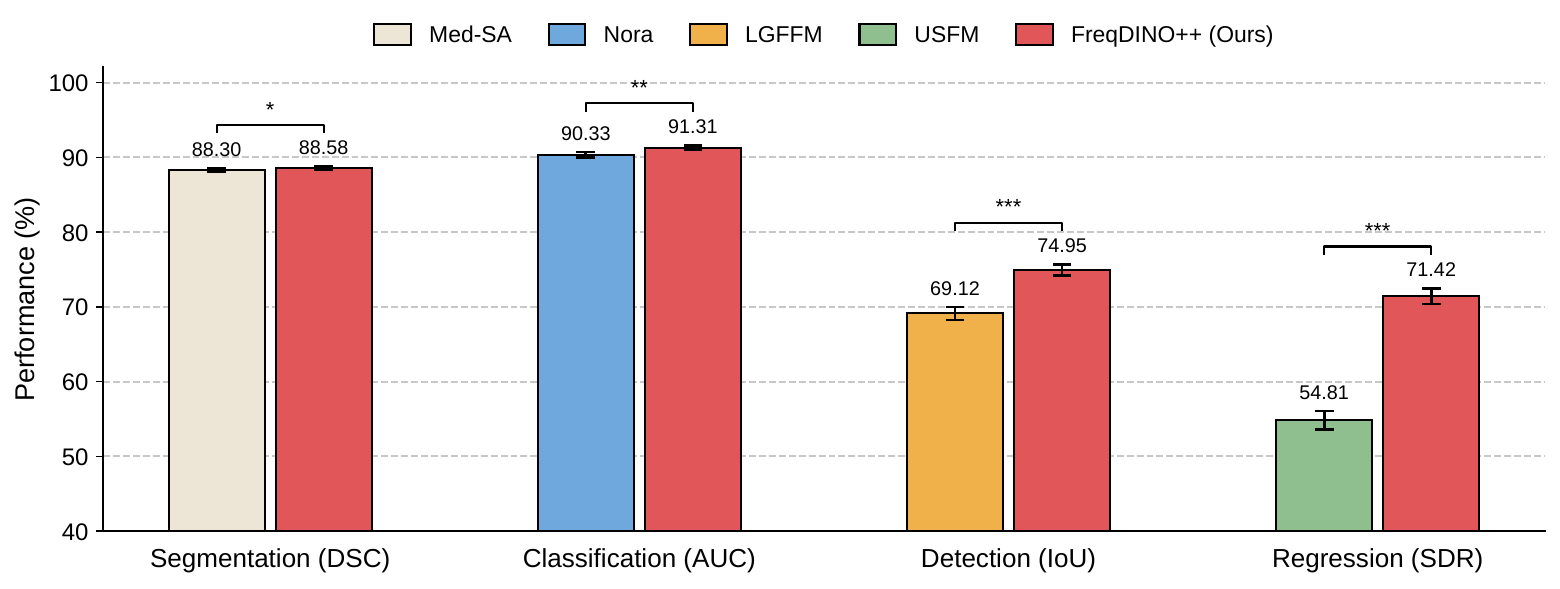}
  \caption{\xq{Statistical analysis of FreqDINO++ across five independent runs on the FMC-UIA multi-task dataset. Each bar reports the mean performance with error bars indicating standard deviation. Statistical significance is assessed by paired $t$-tests ($^{*}$: $p < 0.05$, $^{**}$: $p < 0.01$, $^{***}$: $p < 0.001$).}}
  \label{fig:statis}
\end{figure}

\subsection{Comparison on Multi-Task Ultrasound Analysis}
To comprehensively evaluate the performance of FreqDINO++, we first compare FreqDINO++ with state-of-the-art methods on the FMC-UIA multi-task benchmark. As shown in Table~\ref{tab:multi_task}, classical task-specific methods such as AAU-Net~\cite{chen2022aau} and TransUNet~\cite{chen2024transunet} underperform foundation model-based approaches due to the absence of shared multi-task representations. Among foundation models, SAM-based methods (SAMUS~\cite{lin2024beyond}, UltraSAM~\cite{meyer2025ultrasam}) excel at segmentation but lag on classification, detection, and regression owing to their dense-prediction-oriented designs. Ultrasound-tailored foundation models perform stronger on individual tasks, with USFM~\cite{jiao2024usfm} attaining the best baseline regression (12.33 MRE), LGFFM~\cite{luo2025lgffm} the best detection (38.53\% mAP), and DINOv3-FD~\cite{he2026incentivizing} confirming the transferability of DINOv3 features. Yet all these methods treat each task in isolation, leaving cross-task synergies and frequency-domain cues unexploited.

In contrast, FreqDINO++ delivers the best results across all four task categories simultaneously. It achieves 88.46\% DSC in segmentation with the lowest ASD (6.62) and HD (20.32), reducing them by 0.75 and 1.43 over Med-SA~\cite{wu2025medical}. The gain is more pronounced in classification (91.24\% AUC, 81.57\% F1, 71.99\% MCC), surpassing Nora~\cite{wei2025noise} by 6.64\% F1, and in detection (74.83\% IoU, 47.75\% mAP, 82.84\% AP$_{50}$), outperforming LGFFM by 9.22\% mAP. For regression, it attains 7.74 MRE, 14.21 MAE, and 71.69\% SDR, reducing MRE by 4.59 over USFM. The radar plot in Fig.~\ref{fig:radar} further shows that the polygon of FreqDINO++ consistently encloses those of leading competitors across all 27 subtasks, while the qualitative cases in Fig.~\ref{fig:vis1} corroborate its superiority across heterogeneous tasks.

\begin{table*}[!t]
\centering
\small
\setlength\tabcolsep{5.5pt}
\caption{Comparison with state-of-the-art methods on single-task ultrasound benchmarks.}
\resizebox{\textwidth}{!}{\begin{tabular}{l|ccc|ccc|ccc|ccc}
\hline
\multirow{2}{*}{Methods} & \multicolumn{3}{c|}{BUSI-Seg} & \multicolumn{3}{c|}{BUSI-Class} & \multicolumn{3}{c|}{TN5000} & \multicolumn{3}{c}{FHC} \\
\cline{2-13}
& DSC$\uparrow$ & ASD$\downarrow$ & HD$\downarrow$ & AUC$\uparrow$ & F1$\uparrow$ & MCC$\uparrow$ & IoU$\uparrow$ & mAP$\uparrow$ & AP$_{\rm 50}$$\uparrow$ & MRE$\downarrow$ & MAE$\downarrow$ & SDR$\uparrow$ \\
\hline
AAU-Net \cite{chen2022aau} & 71.31 & 18.94 & 59.51 & 85.63 & 69.93 & 53.12 & 67.71 & 41.74 & 78.63 & 21.19 & 40.79 & 29.58 \\
TransUNet \cite{chen2024transunet} & 72.51 & 21.77 & 73.75 & 95.23 & 84.00 & 75.66 & 68.34 & 40.30 & 79.12 & 15.44 & 34.77 & 37.08 \\
MADGNet \cite{nam2024modality} & 67.32 & 24.17 & 84.07 & 94.79 & 85.54 & 77.16 & 68.50 & 39.44 & 79.32 & 19.38 & 39.86 & 31.67 \\
SAM2 \cite{ravisam} & 75.93 & 20.52 & 79.12 & 80.98 & 62.62 & 42.19 & 50.03 & 15.19 & 44.19 & 26.38 & 46.32 & 22.19 \\
Med-SA \cite{wu2025medical} & 78.01 & 20.09 & 67.06 & 97.74 & 85.39 & 77.38 & 62.37 & 27.11 & 63.97 & 13.75 & 32.07 & 38.85 \\
SAM2-Adapter \cite{chen2025sam2} & 77.81 & 17.39 & 57.82 & 95.30 & 81.91 & 71.00 & 54.43 & 19.71 & 52.12 & 16.34 & 34.65 & 36.67 \\
SAMUS \cite{lin2024beyond} & 60.46 & 34.03 & 111.81 & 91.90 & 85.47 & 74.40 & 68.66 & 35.35 & 78.39 & 14.31 & 31.73 & 41.15 \\
UltraSAM \cite{meyer2025ultrasam}  & 79.97 & 12.78 & 42.98 & 96.16 & 84.35 & 75.61 & 61.25 & 24.42 & 59.24 & 15.20 & 31.67 & 40.42 \\
Nora \cite{wei2025noise} & 76.76 & 15.68 & 61.46 & 97.25 & 87.47 & 77.63 & 60.71 & 25.08 & 62.83 & 14.79 & 30.16 & 42.92 \\
USFM \cite{jiao2024usfm} & 79.35 & 16.50 & 54.30 & 78.15 & 55.32 & 30.46 & 64.52 & 34.26 & 74.46 & 22.20 & 45.45 & 25.52 \\
LGFFM \cite{luo2025lgffm} & 78.72 & 15.42 & 46.82 & 96.64 & 84.90 & 72.64 & 72.43 & 42.89 & 76.01 & 14.63 & 32.20 & 38.12 \\
DINOv3-FD \cite{he2026incentivizing} & 71.98 & 17.20 & 64.18 & 97.82 & 86.66 & 79.15 & 57.23 & 23.63 & 62.24 & 17.23 & 36.52 & 35.73 \\
\xq{UniSeg \cite{ye2023uniseg}} & \xq{73.33} & \xq{20.03} & \xq{67.12} & \xq{96.98} & \xq{84.40} & \xq{75.21} & \xq{66.96} & \xq{34.34} & \xq{74.59} & \xq{15.92} & \xq{35.07} & \xq{38.23} \\
\xq{Hermes \cite{gao2024training}} & \xq{73.32} & \xq{17.03} & \xq{62.48} & \xq{97.04} & \xq{84.69} & \xq{76.69} & \xq{67.57} & \xq{35.65} & \xq{74.55} & \xq{15.68} & \xq{34.82} & \xq{36.67} \\
\xq{TADFormer \cite{baek2025tadformer}} & \xq{75.73} & \xq{16.05} & \xq{63.41} & \xq{96.90} & \xq{85.42} & \xq{77.46} & \xq{66.24} & \xq{32.11} & \xq{73.47} & \xq{15.48} & \xq{34.19} & \xq{40.00} \\
\hline
FreqDINO++ & \textbf{81.79} & \textbf{9.84} & \textbf{31.86} & \textbf{97.91} & \textbf{89.69} & \textbf{83.41} & \textbf{72.69} & \textbf{46.98} & \textbf{85.41} & \textbf{12.37} & \textbf{28.26} & \textbf{47.19} \\
\hline
\end{tabular}}
\label{tab:single_task}
\end{table*}

\xq{Moreover, to assess the robustness and reproducibility of FreqDINO++, we conduct five independent training runs on the FMC-UIA multi-task dataset. As illustrated in Fig. \ref{fig:statis}, FreqDINO++ consistently outperforms the strongest competitor for each task with small standard deviations. Statistical significance is confirmed by paired $t$-tests ($p < 0.05$ for segmentation, $p < 0.01$ for classification, and $p < 0.001$ for detection and regression), and the 95\% confidence intervals of FreqDINO++ do not overlap with the mean performance of the corresponding competitors in classification ([90.93, 91.69] vs. 90.33), detection ([74.07, 75.83] vs. 69.12), and regression ([70.12, 72.72] vs. 54.81), further confirming the statistical reliability of the reported improvements.} These results confirm the effectiveness of our frequency-guided adaptation and multi-task routing for universal ultrasound analysis.

\begin{figure*}[!t]
  \centering
  \includegraphics[width=1\linewidth]{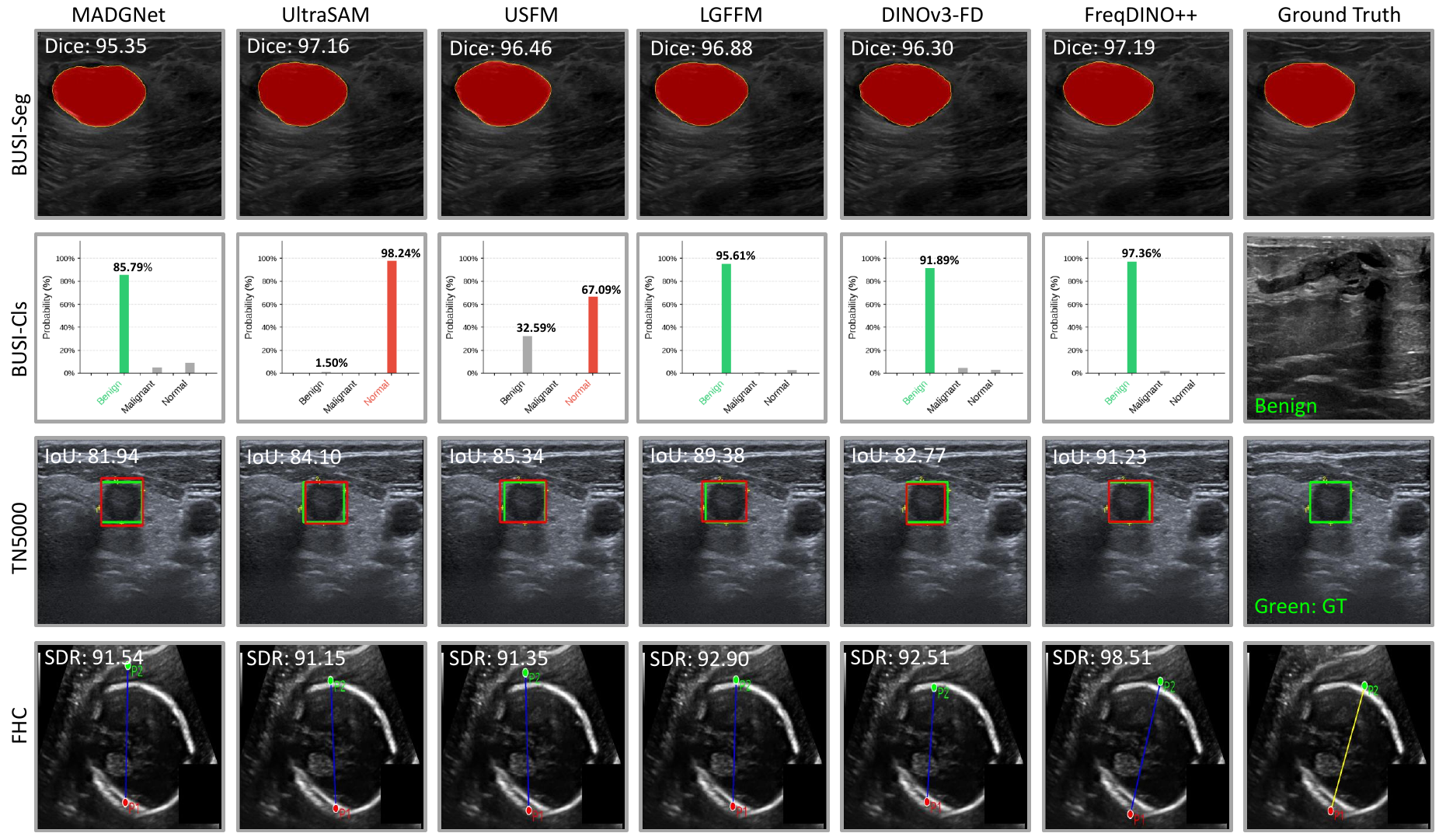}
    \caption{Visualization of qualitative results on four benchmarks. Our FreqDINO++ consistently achieves the best results across all four single-task settings, with more accurate masks, correct categories, tighter detection boxes, and more precise biometric landmarks than competing methods.}
  \label{fig:vis2}
\end{figure*}

\subsection{Comparison on Single-Task Benchmarks}
We further evaluate FreqDINO++ on four external single-task benchmarks to assess its transferability beyond the multi-task training setting. As reported in Table \ref{tab:single_task}, classical task-specific methods exhibit limited performance, with AAU-Net \cite{chen2022aau} and MADGNet \cite{nam2024modality} obtaining only 71.31\% and 67.32\% DSC on BUSI segmentation. Foundation model-based approaches achieve stronger task-specific results: UltraSAM \cite{meyer2025ultrasam} reaches 79.97\% DSC on BUSI segmentation, DINOv3-FD \cite{he2026incentivizing} attains 97.82\% AUC on BUSI classification, LGFFM \cite{luo2025lgffm} obtains 42.89\% mAP on TN5000 detection, and Nora \cite{wei2025noise} achieves 42.92\% SDR on FHC regression. However, no single baseline performs consistently across all four tasks.

In contrast, our FreqDINO++ achieves state-of-the-art performance across all four single-task benchmarks simultaneously. On BUSI segmentation, FreqDINO++ attains 81.79\% DSC and reduces HD to 31.86, surpassing UltraSAM by 1.82\% in DSC. For BUSI classification, it obtains 97.91\% AUC and 89.69\% F1, outperforming Nora by 2.22\% in F1. On TN5000 detection, FreqDINO++ reaches 46.98\% mAP and 85.41\% AP$_{\rm 50}$, improving over LGFFM by 4.09\% in mAP. For FHC regression, it achieves the lowest MRE of 12.37 and the highest SDR of 47.19\%, surpassing Nora by 4.27\%. Qualitative comparisons of representative cases across the four task categories are further illustrated in Fig. \ref{fig:vis2}. These results confirm that, despite being trained within a unified multi-task framework, the proposed FreqDINO++ framework delivers consistently superior performance across diverse external benchmarks and outperforms task-specific methods.

\subsection{Ablation Study}
To investigate the effectiveness of the MR-Adapter, F$^2$-Enhancer, and TC-Decoder, we conduct a comprehensive ablation study on the FMC-UIA multi-task dataset, as illustrated in Table \ref{tab:ablation}. The frozen DINOv3 backbone with vanilla linear decoders (1$^{st}$ row) serves as the ablation baseline. By separately introducing the MR-Adapter (2$^{nd}$ row), F$^2$-Enhancer (3$^{rd}$ row), and TC-Decoder (4$^{th}$ row), segmentation DSC is consistently improved over the baseline by 0.76\%, 1.38\%, and 0.64\%, respectively. Among them, the F$^2$-Enhancer yields the largest segmentation gain (1.38\% DSC) and regression refinement (1.18 MRE reduction), the TC-Decoder contributes the most to detection (1.34\% mAP), and the MR-Adapter brings the largest regression SDR improvement (4.05\%).

We further examine pairwise combinations. By comparing the 6$^{th}$ and 7$^{th}$ rows with the corresponding single-module configurations, the TC-Decoder consistently amplifies the benefits of both the MR-Adapter and F$^2$-Enhancer, particularly boosting detection mAP by 3.90\% and regression SDR by 6.87\%. Building upon these complementary effects, our complete FreqDINO++ (8$^{th}$ row) integrates all three modules to achieve the best performance across all tasks, with overall gains of 2.97\% DSC, 12.03\% F1, 5.93\% mAP, and 12.44\% SDR over the baseline. These comprehensive ablation experiments validate that the tailored MR-Adapter, F$^2$-Enhancer, and TC-Decoder collectively contribute to the superior performance of FreqDINO++ by enabling parameter-efficient task-conditioned adaptation, capturing multi-scale frequency characteristics, and coordinating dense and global prediction tasks through token-level collaboration.

\begin{table*}[!t]
\centering
\small
\setlength\tabcolsep{3.5pt}
\caption{Ablation study of FreqDINO++ on the FMC-UIA multi-task dataset.}
\resizebox{\textwidth}{!}{\begin{tabular}{ccc|ccc|ccc|ccc|ccc}
\hline
\multirow{2}{*}{MR-Adapter} & \multirow{2}{*}{F$^2$-Enhancer} & \multirow{2}{*}{TC-Decoder} & \multicolumn{3}{c|}{Segmentation} & \multicolumn{3}{c|}{Classification} & \multicolumn{3}{c|}{Detection} & \multicolumn{3}{c}{Regression} \\
\cline{4-15}
& & & DSC$\uparrow$ & ASD$\downarrow$ & HD$\downarrow$ & AUC$\uparrow$ & F1$\uparrow$ & MCC$\uparrow$ & IoU$\uparrow$ & mAP$\uparrow$ & AP$_{\rm 50}$$\uparrow$ & MRE$\downarrow$ & MAE$\downarrow$ & SDR$\uparrow$ \\
\hline
 & & & 85.49 & 8.39 & 26.33 & 89.94 & 69.54 & 60.75 & 70.11 & 41.82 & 77.33 & 10.24 & 19.47 & 59.25 \\
\checkmark & & & 86.25 & 7.74 & 23.46 & 89.57 & 73.07 & 62.76 & 71.12 & 40.26 & 82.14 & 9.52 & 17.38 & 63.30 \\
 & \checkmark & & 86.87 & 7.64 & 24.43 & 90.14 & 74.42 & 63.25 & 72.05 & 40.83 & 81.53 & 9.06 & 17.25 & 62.11 \\
 & & \checkmark & 86.13 & 7.92 & 23.58 & 90.01 & 73.28 & 62.61 & 71.81 & 43.16 & 82.22 & 9.71 & 18.20 & 61.79 \\
\checkmark & \checkmark & & 87.16 & 7.31 & 22.03 & 90.71 & 76.20 & 65.40 & 72.53 & 43.47 & 83.03 & 8.51 & 15.76 & 64.86 \\
\checkmark & & \checkmark & 87.06 & 7.52 & 22.26 & 90.80 & 78.30 & 66.80 & 73.25 & 44.16 & 82.63 & 8.61 & 15.22 & 69.56 \\
 & \checkmark & \checkmark & 87.79 & 6.97 & 21.73 & 90.63 & 78.64 & 68.15 & 72.77 & 43.47 & 79.99 & 8.48 & 15.76 & 68.98 \\
\checkmark & \checkmark & \checkmark & \textbf{88.46} & \textbf{6.62} & \textbf{20.32} & \textbf{91.24} & \textbf{81.57} & \textbf{71.99} & \textbf{74.83} & \textbf{47.75} & \textbf{82.84} & \textbf{7.74} & \textbf{14.21} & \textbf{71.69} \\
\hline
\end{tabular}}
\label{tab:ablation}
\end{table*}

\begin{table*}[!t]
\centering
\small
\setlength\tabcolsep{5.5pt}
\caption{Effectiveness of Frequency-aware Feature Enhancer with different frequency band combinations on the FMC-UIA dataset.}
\resizebox{\textwidth}{!}{\begin{tabular}{ccc|ccc|ccc|ccc|ccc}
\hline
\multirow{2}{*}{LF} & \multirow{2}{*}{MF} & \multirow{2}{*}{HF} & \multicolumn{3}{c|}{Segmentation} & \multicolumn{3}{c|}{Classification} & \multicolumn{3}{c|}{Detection} & \multicolumn{3}{c}{Regression} \\
\cline{4-15}
& & & DSC$\uparrow$ & ASD$\downarrow$ & HD$\downarrow$ & AUC$\uparrow$ & F1$\uparrow$ & MCC$\uparrow$ & IoU$\uparrow$ & mAP$\uparrow$ & AP$_{\rm 50}$$\uparrow$ & MRE$\downarrow$ & MAE$\downarrow$ & SDR$\uparrow$ \\
\hline
 & & & 87.15 & 7.44 & 21.98 & 90.21 & 77.17 & 65.97 & 73.07 & 44.62 & 81.49 & 8.44 & 15.47 & 67.59 \\
\checkmark & & & 87.47 & 7.22 & 21.86 & 90.51 & 77.34 & 67.88 & 73.58 & 46.07 & 82.70 & 8.39 & 15.01 & 68.24 \\
\checkmark & \checkmark & & 88.29 & 6.77 & 20.48 & 91.03 & 81.00 & 70.70 & 74.40 & 46.06 & 82.52 & 8.23 & 15.23 & 68.95 \\
\checkmark & \checkmark & \checkmark & \textbf{88.46} & \textbf{6.62} & \textbf{20.32} & \textbf{91.24} & \textbf{81.57} & \textbf{71.99} & \textbf{74.83} & \textbf{47.75} & \textbf{82.84} & \textbf{7.74} & \textbf{14.21} & \textbf{71.69} \\

\hline
\end{tabular}}
\label{tab:freq_ablation}
\end{table*}

\begin{figure}[!t]
  \centering
  \includegraphics[width=1\linewidth]{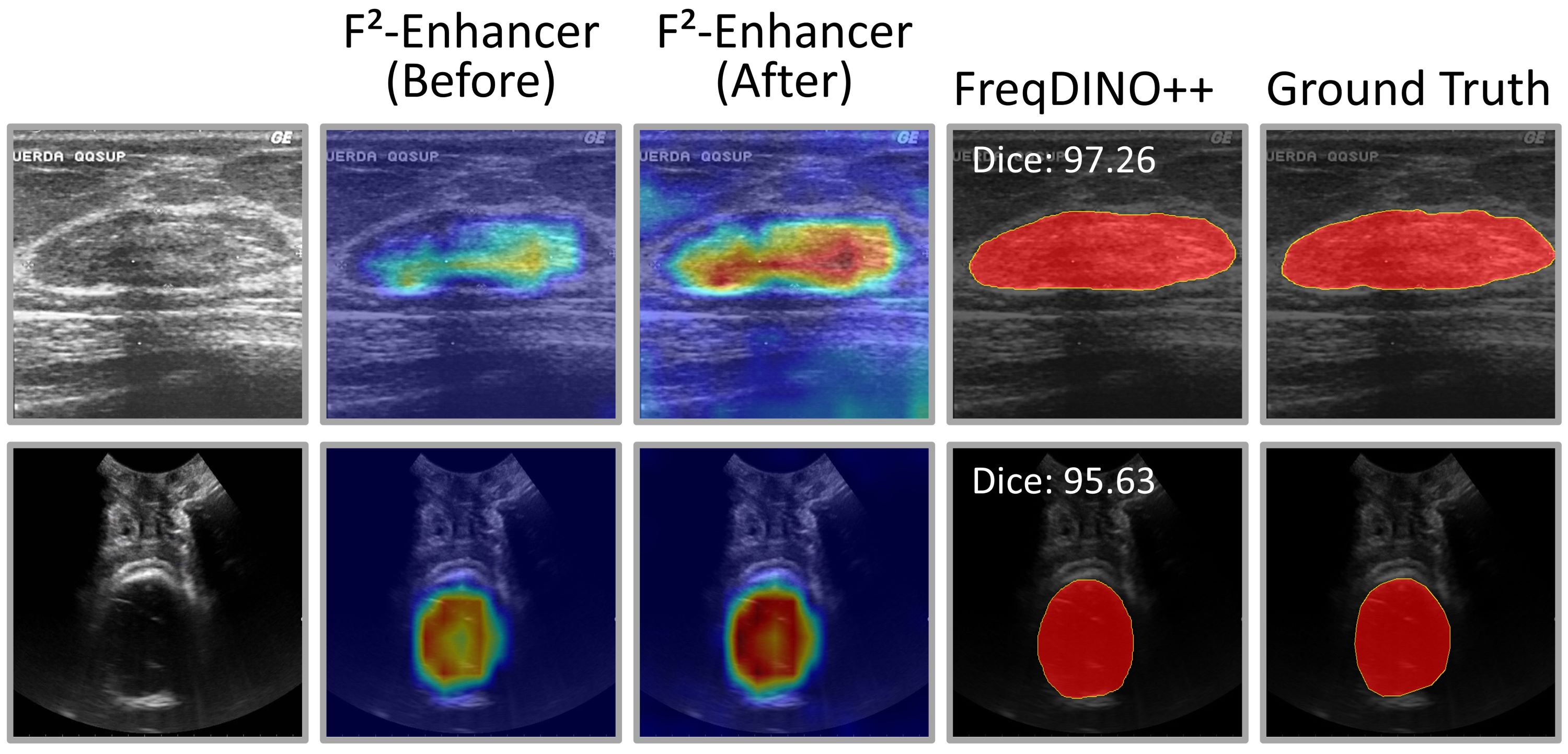}
  \caption{\xq{Visualization of attention maps before and after the F$^2$-Enhancer on the FMC-UIA dataset. Our FreqDINO++ with the F$^2$-Enhancer produces more focused attention on target regions with sharper boundary delineation and accurate segmentation predictions.}}
  \label{fig:attention}
\end{figure}

\begin{table}[!t]
\centering
\small
\setlength\tabcolsep{11pt}
\caption{Comparison on generalization capability on the unseen TN3K segmentation dataset.}
\resizebox{\columnwidth}{!}{\begin{tabular}{l|ccc}
\hline
Methods & DSC$\uparrow$ & ASD$\downarrow$ & HD$\downarrow$ \\
\hline
AAU-Net \cite{chen2022aau} & 68.49 & 20.82 & 66.89 \\
TransUNet \cite{chen2024transunet} & 76.68 & 15.48 & 50.58 \\
MADGNet \cite{nam2024modality} & 71.69 & 18.38 & 61.41 \\
SAM2 \cite{ravisam} & 76.25 & 15.25 & 47.82 \\
Med-SA \cite{wu2025medical} & 80.18 & 13.43 & 40.18 \\
SAM2-Adapter \cite{chen2025sam2} & 75.75 & 16.12 & 54.06 \\
SAMUS \cite{lin2024beyond} & 79.84 & 13.31 & 45.61 \\
UltraSAM \cite{meyer2025ultrasam} & 79.93 & 12.62 & 39.43 \\
Nora \cite{wei2025noise} & 80.81 & 12.13 & 39.14 \\
USFM \cite{jiao2024usfm} & 77.70 & 12.55 & 41.63 \\
LGFFM \cite{luo2025lgffm} & 80.34 & 12.76 & 42.83 \\
DINOv3-FD \cite{he2026incentivizing} & 74.42 & 16.94 & 62.18 \\
\xq{UniSeg \cite{ye2023uniseg}} & \xq{80.16} & \xq{12.08} & \xq{39.01} \\
\xq{Hermes \cite{gao2024training}} & \xq{80.43} & \xq{12.03} & \xq{39.39} \\
\xq{TADFormer \cite{baek2025tadformer}} & \xq{80.61} & \xq{12.26} & \xq{38.94} \\
\hline
FreqDINO++ & \textbf{82.86} & \textbf{10.71} & \textbf{37.62} \\
\hline
\end{tabular}}
\label{tab:generalization}
\end{table}

\subsection{Effectiveness of Frequency-aware Feature Enhancer}
To evaluate the contribution of each frequency band in the F$^2$-Enhancer, we progressively incorporate different frequency components through cascaded wavelet decomposition, as shown in Table~\ref{tab:freq_ablation}. Without any frequency decomposition, the framework operates entirely in the spatial domain and achieves baseline performance across all four tasks (1$^{st}$ row). Introducing the low-frequency band alone (2$^{nd}$ row) yields consistent gains across all tasks, with 0.30\% AUC improvement and 1.91\% MCC gain in classification, as the LF band captures global anatomical structures that benefit holistic prediction. Sequentially adding the mid-frequency band (3$^{rd}$ row) further boosts classification with a 3.66\% F1 gain, confirming that tissue texture patterns encoded in the MF band are critical for diagnostic categorization. The complete three-band integration (4$^{th}$ row), incorporating the high-frequency band for boundary details, achieves optimal performance with 88.46\% DSC and 6.62 ASD in segmentation, representing total gains of 1.31\% DSC, 4.40\% F1, 3.13\% mAP, and 4.10\% SDR over the spatial-only baseline. The progressive improvements across band combinations confirm that the LF, MF, and HF bands play distinct yet synergistic roles, jointly aligning with the structural, textural, and boundary-related demands of heterogeneous ultrasound tasks.

\xq{To further provide direct evidence that the F$^2$-Enhancer effectively improves feature representations, we visualize the attention maps before and after the F$^2$-Enhancer on the FMC-UIA dataset. As illustrated in Fig. \ref{fig:attention}, the attention maps before the F$^2$-Enhancer exhibit diffuse activation patterns that spread across both the target lesion and surrounding background regions. After applying the F$^2$-Enhancer, the attention maps become significantly more focused and concentrated on the target anatomical regions with sharper boundary delineation and suppressed background activation, demonstrating that the cascaded cross-band attention effectively aggregates complementary frequency information to enhance lesion-specific feature representations.}

\begin{table}[!t]
\centering
\small
\setlength\tabcolsep{11pt}
\caption{\xq{Comparison on generalization capability on the unseen BrEaST classification dataset.}}
\resizebox{\columnwidth}{!}{\begin{tabular}{l|ccc}
\hline
Methods & AUC$\uparrow$ & F1$\uparrow$ & MCC$\uparrow$ \\
\hline
AAU-Net \cite{chen2022aau} & 59.17 & 30.10 & 9.71 \\
TransUNet \cite{chen2024transunet} & 74.88 & 41.01 & 28.80 \\
MADGNet \cite{nam2024modality} & 71.98 & 19.32 & 13.99 \\
SAM2 \cite{ravisam} & 60.18 & 27.34 & 18.27 \\
Med-SA \cite{wu2025medical} & 83.15 & 44.16 & 33.06 \\
SAM2-Adapter \cite{chen2025sam2} & 71.02 & 20.22 & 14.26 \\
SAMUS \cite{lin2024beyond} & 78.79 & 36.59 & 25.16 \\
UltraSAM \cite{meyer2025ultrasam} & 83.99 & 48.71 & 35.91 \\
Nora \cite{wei2025noise} & 83.49 & 45.09 & 33.79 \\
USFM \cite{jiao2024usfm} & 67.36 & 27.69 & 15.45 \\
LGFFM \cite{luo2025lgffm} & 84.11 & 35.46 & 25.62 \\
DINOv3-FD \cite{he2026incentivizing} & 66.67 & 31.54 & 15.41 \\
UniSeg \cite{ye2023uniseg} & 69.55 & 32.84 & 15.69 \\
Hermes \cite{gao2024training} & 71.70 & 31.32 & 13.63 \\
TADFormer \cite{baek2025tadformer} & 71.88 & 34.11 & 15.87 \\
\hline
FreqDINO++ & \textbf{84.59} & \textbf{51.23} & \textbf{41.84} \\
\hline
\end{tabular}}
\label{tab:generalization_BrEaST}
\end{table}

\begin{table*}[!t]
\centering
\small
\setlength\tabcolsep{11pt}
\caption{\xq{Comparison of computational efficiency with state-of-the-art foundation models.}}
\resizebox{\textwidth}{!}{\begin{tabular}{l|ccccccc}
\hline
\multirow{2}{*}{Methods} & \#Total & \#Trainable & Adapter & \multirow{2}{*}{FLOPs (G)$\downarrow$} & GPU Mem & Inference & Training \\
& Params (M)$\downarrow$ & Params (M)$\downarrow$ & Ratio (\%)$\downarrow$ & & (GB)$\downarrow$ & Latency (ms)$\downarrow$ & Time (h)$\downarrow$ \\
\hline
SAM2 \cite{ravisam} & 901.52 & 50.68 & 5.62 & 407.32 & 3.60 & 82.12 & 23.28 \\
Med-SA \cite{wu2025medical} & 1287.72 & 70.32 & 5.46 & 1004.00 & 5.20 & 106.96 & 124.76 \\
SAM2-Adapter \cite{chen2025sam2} & 906.56 & 55.76 & 6.15 & 408.68 & 3.64 & 92.16 & 24.16 \\
SAMUS \cite{lin2024beyond} & 536.92 & 162.24 & 30.22 & 11670.20 & 2.84 & 46.64 & 138.28 \\
UltraSAM \cite{meyer2025ultrasam} & 366.76 & 366.76 & 100.00 & 258.32 & 2.28 & 62.56 & 51.48 \\
Nora \cite{wei2025noise} & 553.56 & 206.84 & 37.37 & 266.76 & 2.28 & 82.00 & 41.20 \\
USFM \cite{jiao2024usfm} & 1151.52 & 1151.52 & 100.00 & 370.04 & 5.56 & 18.28 & 19.64 \\
LGFFM \cite{luo2025lgffm} & 1111.92 & 63.32 & 6.94 & 512.64 & 4.28 & 121.84 & 134.72 \\
DINOv3-FD \cite{he2026incentivizing} & 1246.36 & \textbf{34.04} & \textbf{2.73} & 788.96 & 4.88 & 82.72 & 29.00 \\
\hline
FreqDINO++ & \textbf{344.36} & 41.28 & 11.99 & \textbf{205.40} & \textbf{2.72} & \textbf{18.14} & \textbf{11.73} \\
\hline
\end{tabular}}
\label{tab:efficiency}
\end{table*}

\subsection{Comparison on Generalization Capability}
To further validate the generalization capability of FreqDINO++, we evaluate all methods on the unseen TN3K thyroid nodule segmentation dataset, where all models directly apply the weights trained on the FMC-UIA multi-task dataset without any fine-tuning. As shown in Table \ref{tab:generalization}, classical methods such as AAU-Net \cite{chen2022aau} and MADGNet \cite{nam2024modality} suffer substantial performance degradation on this unseen dataset, achieving only 68.49\% and 71.69\% DSC, respectively, as their task-specific architectures lack the transferable representations needed for cross-dataset generalization. Foundation model-based approaches demonstrate comparatively stronger robustness, with Nora \cite{wei2025noise} and LGFFM \cite{luo2025lgffm} attaining 80.81\% and 80.34\% DSC, benefiting from their large-scale pre-trained features. In contrast, our FreqDINO++ achieves the best generalization performance with 82.86\% DSC, 10.71 ASD, and 37.62 HD, surpassing the best baseline Nora by 2.05\% in DSC and reducing HD by 1.52. 

\xq{To further validate the generalization capability beyond segmentation, we evaluate all methods on the unseen BrEaST \cite{pawlowska2024curated} breast ultrasound classification dataset without any fine-tuning. As shown in Table \ref{tab:generalization_BrEaST}, FreqDINO++ achieves the best generalization performance with 84.59\% AUC, 51.23\% F1, and 41.84\% MCC, outperforming the second-best UltraSAM by 0.60\% AUC, 2.52\% F1, and 5.93\% MCC, confirming that FreqDINO++ generalizes well across different task types on unseen datasets.} These results demonstrate that the frequency-guided adaptation and multi-task routing mechanisms in FreqDINO++ not only enhance in-domain performance but also foster more transferable representations on unseen ultrasound datasets, confirming its significant potential for deployment in real-world clinical scenarios.

\subsection{Computational Efficiency}
\xq{To assess the practical applicability of FreqDINO++, we compare the computational efficiency with state-of-the-art foundation models in Table \ref{tab:efficiency}. FreqDINO++ achieves the lowest computational cost with 205.40G FLOPs and 18.14 ms inference latency per image, corresponding to approximately 55 FPS that well exceeds the typical ultrasound frame rate of 20--30 FPS for real-time clinical deployment. In terms of model complexity, FreqDINO++ requires only 344.36M total parameters with 41.28M trainable parameters and consumes only 2.72 GB GPU memory. The training process completes in 11.73 hours, which is the fastest among all compared methods. The significant computational efficiency stems from our parameter-efficient adaptation paradigm, which freezes the DINOv3 backbone and introduces only lightweight MR-Adapter modules for task-conditioned routing.}

\section{Conclusion}
In this work, we propose FreqDINO++, a frequency-guided multi-task routing framework for universal ultrasound analysis to address the challenges of adapting vision foundation models to heterogeneous ultrasound tasks. Built on a frozen DINOv3 backbone, FreqDINO++ integrates a MR-Adapter to efficiently balance task-common and task-specific knowledge through deterministic task-conditioned routing, and a F$^2$-Enhancer that decomposes features into multiple frequency bands via cascaded wavelet transforms and progressively aggregates them to capture rich frequency information inherent in ultrasound images. Additionally, a TC-Decoder leverages the collaborative interaction between local and global token representations to jointly serve segmentation, classification, detection, and regression tasks within a unified dual-branch architecture. Extensive experiments on diverse datasets show that FreqDINO++ consistently surpasses state-of-the-art methods across all four task categories, \textit{i.e.}, segmentation, classification, detection, and regression settings, while also exhibiting strong generalization capability on unseen datasets. 

\balance

\bibliographystyle{IEEEtran}
\bibliography{references}

\end{document}